\documentclass[conference]{IEEEtran}
\IEEEoverridecommandlockouts
\usepackage{cite}
\usepackage{xcolor}
\usepackage{amsmath,amssymb,amsfonts}
\usepackage{algorithmic}
\usepackage{graphicx}
\usepackage{array}
\usepackage{textcomp}
\usepackage{stfloats}
\usepackage{url}
\usepackage{verbatim}
\usepackage{graphicx}
\usepackage[pagebackref,breaklinks,citecolor=citecolor,colorlinks]{hyperref}

\usepackage{booktabs}
\usepackage{graphicx}
\def\BibTeX{{\rm B\kern-.05em{\sc i\kern-.025em b}\kern-.08em
    T\kern-.1667em\lower.7ex\hbox{E}\kern-.125emX}}

\definecolor{dg}{rgb}{0,0.694,0.298}
\definecolor{purple}{rgb}{0.4,0.176,0.569}
\definecolor{Gray}{gray}{0.6}
\definecolor{citecolor}{RGB}{65,105,225}
\begin{document}

\title{Zero-shot Face Editing via ID-Attribute Decoupled Inversion}

\author{
  Yang Hou$^{*}$, Minggu Wang, Jianjun Zhao\\
  Graduate School and Faculty of Information Science and Electrical Engineering,\\
  Kyushu University, Fukuoka, Japan\\
  \{hou.yang.549, wang.minggu.065\}@s.kyushu-u.ac.jp, zhao@ait.kyushu-u.ac.jp
}

\maketitle

\begin{abstract}
Recent advancements in text-guided diffusion models have shown promise for general image editing via inversion techniques, but often struggle to maintain ID and structural consistency in real face editing tasks. 
To address this limitation, we propose a zero-shot face editing method based on ID-Attribute Decoupled Inversion. Specifically, we decompose the face representation into ID and attribute features, using them as joint conditions to guide both the inversion and the reverse diffusion processes. This allows independent control over ID and attributes, ensuring strong ID preservation and structural consistency while enabling precise facial attribute manipulation. 
Our method supports a wide range of complex multi-attribute face editing tasks using only text prompts, without requiring region-specific input, and operates at a speed comparable to DDIM inversion. Comprehensive experiments demonstrate its practicality and effectiveness.

\end{abstract}


\section{Introduction}
\graphicspath{{images/}}
\begin{figure}[t]
  \centering
  \includegraphics[width=\linewidth]{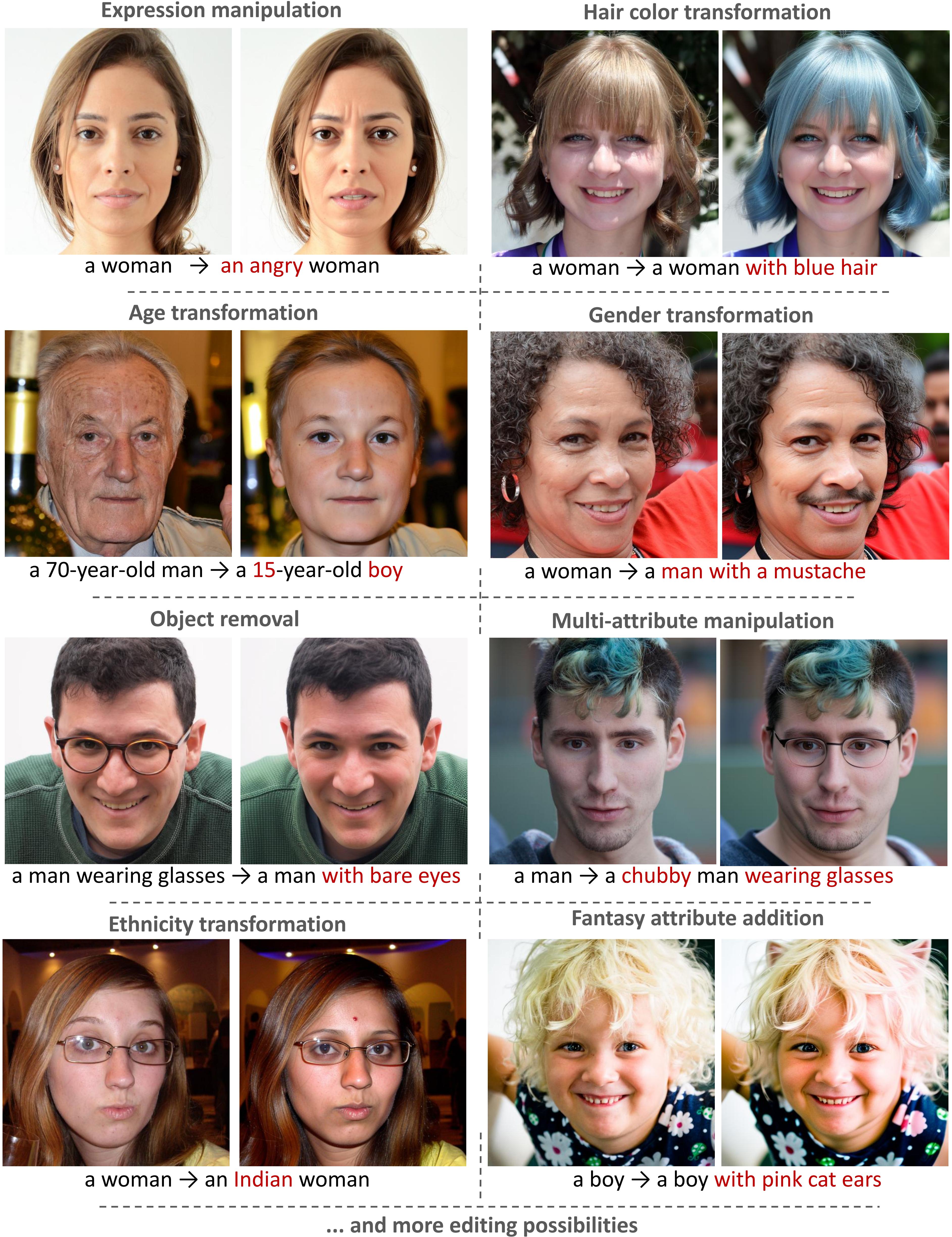}
  \caption{ In each pair of images, the left shows the original input image with its corresponding text description displayed below. The right shows the edited image, with the modified text description displayed below it. we edit the face image based on the modified text description. (Zoom in to see details)}
  \label{fig:fig1}
\end{figure}

Face editing poses greater challenges than general image editing, as it demands the modification of complex and intertwined facial attributes while strictly preserving identity and structural consistency to ensure the face remains recognizable and retains its original structure.

Currently, GAN-based face editing methods have been extensively studied and have achieved promising results \cite{shen2020interpreting}, In contrast, diffusion models, despite their recent breakthroughs in image generation and general image editing \cite{stablediffusion,Glide,ramesh2022hierarchical}, remain relatively underexplored for face editing tasks.

For diffusion models, inversion-based methods are one of the mainstream approaches for image editing, typically involving two steps: first, the image is inverted into latent space as initial latent code using an inversion technique; next, this initial latent code serves as the starting point for the reverse diffusion process to modify specific content under new conditions. This two-step approach provides a flexible framework for various editing tasks, enabling feature disentanglement in the latent space and allowing independent control over specific features.  However, it faces additional challenges when applied to text-guided diffusion models.

In text-guided diffusion models \cite{stablediffusion}, while the text condition provides a more flexible way to control target features, it also complicates feature disentanglement in the latent space, making precise control over specific features more challenging. Additionally, classifier-free guidance (CFG) \cite{ho2022classifier} often leads to edited images that deviate significantly from the original, making it difficult to maintain structural consistency. To address this challenge, some works have explored the use of references to improve structural consistency, such as PnP \cite{plug-and-play} and Pix2pix-zero \cite{pix2pixzero}. These methods typically use a DDIM \cite{DDIM} reconstruction trajectory as a reference to guide the reverse diffusion process by providing structural constraints (e.g., attention maps or latent codes in each of the diffusion steps).

\begin{figure*}[ht]
    \centering
    \includegraphics[width=\textwidth]{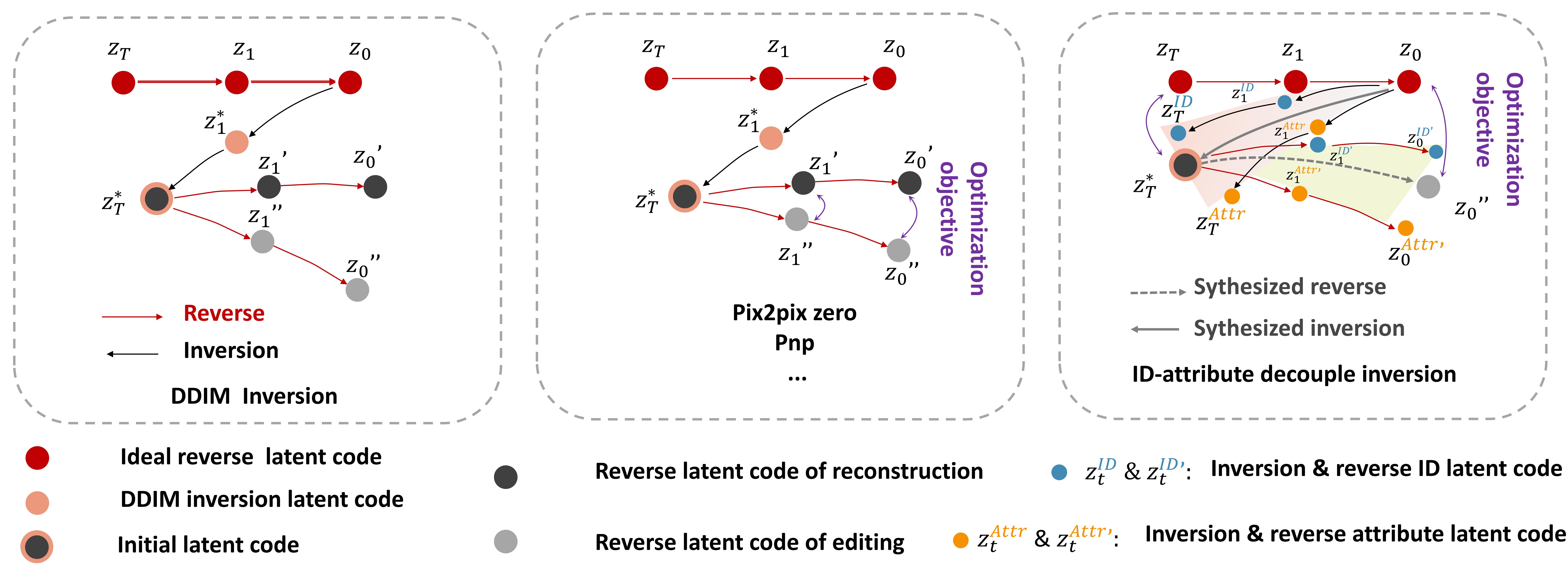}
    \caption{
The left diagram illustrates a \( T \)-step DDIM inversion and reverse diffusion process, where \( z_{T} \rightarrow z_{0} \) represents the ideal reverse diffusion denoising trajectory. \( z_{0} \rightarrow z^*_{T} \) denotes the DDIM inversion trajectory guided by the text condition \( P \), yielding \( z^*_{T} \) as an approximation of \( z_{T} \). \( z^*_T \rightarrow z'_0 \) is the reconstruction trajectory under the guidance of condition \( P \), while \( z^*_T \rightarrow z''_0 \) represents the reverse diffusion process trajectory guided by a new condition \( P_n \), resulting in \( z''_0 \) deviating significantly from \( z_0 \). The middle diagram illustrates existing inversion-based image editing method, which typically use the reconstruction trajectory \( z^*_T \rightarrow z'_0 \) as a reference to optimize \( z^*_T \rightarrow z''_0 \). The right diagram illustrates our method, which uses both ID features and facial attributes as joint conditions to guide the inversion and reverse diffusion processes. Under the guidance of these two conditions, the inversion yields a synthesized \( z^*_T \) that is closer to the ideal initial latent code \( z_{T} \). The reverse diffusion process then starts from \( z^*_T \) and results in the synthesized output \( z''_0 \), which is pulled towards \( z_0 \) under the constraint of the input conditions.}
    \label{fig:principle}
\end{figure*}

While these methods are effective for general image editing, they encounter two key limitations in face editing tasks: (1) It does not account for the specificity of face ID features, making it difficult to maintain ID consistency in the edited image. (2) The inversion process is guided by text, which lacks precision in capturing fine-grained facial details, leading to suboptimal initial latent code for the reverse diffusion process and ultimately affecting structural consistency.

To address these limitations, we propose a face editing method via ID-Attribute Decoupled Inversion. As shown in Figure \ref{fig:fig1}, our method can handle complex face editing tasks while maintaining ID and structural consistency. The principle is illustrated in the rightmost diagram of Figure \ref{fig:principle}, with the middle diagram showing the principle of existing inversion-based image editing methods for comparison. The core of our method is enabling the model to decouple ID and other attribute features, allowing for independent control of each. Specifically, we decompose the facial representation into ID features, represented by the entire face image embedding, and attribute features, represented by text embedding, and fine-tune a pre-trained text-guided diffusion model using these two conditions jointly. For training, we build a dataset of face-attribute descriptions, consisting of 69,900 facial attribute descriptions paired with corresponding face images. For editing, we first leverage both conditions to jointly guide the inversion process and obtain an initial latent code. We then use image embedding and the modified text embedding to jointly guide the reverse diffusion process. In this process, the entire face image embedding serves as a fine-grained condition to preserve ID and as a robust constraint to maintain structural consistency, while the text embedding acts as a flexible condition for attribute disentanglement and enables modifications. 


Our method relies on text descriptions for editing, enabling it to localize and modify target attributes without requiring any region-specific inputs. Additionally, it avoids time-consuming optimization for alignment with a reference, achieving an editing speed comparable to DDIM inversion.

Our contributions can be summarized as follows:
\begin{itemize}
\item We propose a zero-shot face editing method based on ID-attribute decoupled inversion, capable of handling a wide range of complex face editing tasks while maintaining ID and structural consistency.

\item We provide the insight that high-dimensional, structured image embeddings can serve as a fine-grained condition to obtain a precise initial latent code and as a robust constraint to align reverse diffusion process trajectory with inversion trajectory. thereby maintaining overall structural consistency.

\item We conduct a comprehensive comparison with state-of-the-art face editing methods. Experimental results demonstrate that our method outperforms others in ID and structural preservation, flexibility, and editing quality.

\end{itemize}
\section{Background and Motivation}
Diffusion models have shown promise in general image editing. For instance, SDEdit \cite{sdedit} adds noise to the entire image and denoises from a specified step to achieve global edits, while later work incorporates masks for localized edits. To further refine editing accuracy, inversion-based approaches have emerged, mapping real images back into a latent space (often via DDIM inversion \cite{DDIM} or its variants) to better disentangle features. These methods focus on structural consistency with the original image and faster editing, but are designed primarily for general image editing. Although many of these methods demonstrate their effectiveness on face editing tasks, they rarely account for ID consistency, which is crucial for face editing.

Meanwhile, most face editing methods still rely on GANs. Only a few diffusion-based face editing methods, such as Diffusion Autoencoders \cite{DiffusionAutoencoders} and Collaborative Diffusion \cite{collaborative}, have been proposed. However, they struggle to maintain fine-grained facial details and ID consistency, limiting their practical applicability (a detailed discussion of related work is provided in the Appendix). Alternatively, face-driven image generation methods, such as IP-adapter FaceID \cite{IP-adapter}, InstantID \cite{InstantID}, and PhotoMaker \cite{photomaker}, focus on preserving ID by creating personalized images that resemble a given input face. However, these approaches are primarily designed for image generation rather than attribute-level editing and often result in significant structural or detail shifts from the input image.

Motivated by these works, our method builds upon the inversion-based image editing framework and incorporates ID-preserving strategies inspired by face-driven image generation approaches, aiming to achieve precise face editing while ensuring both ID and structural consistency.

\section{Methods}
Our objective is to modify the attributes of an input face image \( I \) based on text prompts, transforming it into a target image \( I^* \). The face editing process begins by inverting the input image \( I \) into the latent space under the guidance of initial prompt \( P \), producing the corresponding initial latent code. We then modify the semantics of \( P \) by replacing, adding, or removing words to create a new text description \( P_n \). The modified prompt \( P_n \) guides the reverse diffusion process to generate the edited image. The task requires that the edited face image retains the ID and preserves the structure of the \( I \) while achieving the desired attribute changes.

\textbf{ID and attribute representation.} For a face image, ID features encompass the geometric structure, distinctive textures, and the specific arrangement and proportions of facial details. To accurately represent the ID features of a face image, we utilize the entire face image embedding as the ID feature. we use a pre-trained CLIP vision model \cite{clip} as the image encoder, and employ a projection network to map the entire face image into an embedding. This high-dimensional, structured embedding not only provides a unique representation of facial ID features but also captures fine-grained information from the entire image.

Attribute features capture non-ID characteristics of the face, such as expressions, age, and gender, which can vary for the same individual without altering their ID. To enable flexible modification of these attributes, we represent them using text descriptions.

\textbf{Training.} We train the diffusion model using both the face image's text description and its embedding as joint conditions. The training has two main objectives: (1)training the model to map the face image embedding back to a face image, thereby allowing it to use the image embedding to guide the reverse diffusion process; and (2) aligning the latent codes guided by the image embedding and text embedding to the same distribution, thereby achieving attribute feature-text alignment and disentanglement (i.e., minimizing the distance between \( z_{t}^{\textcolor{cyan}{ID}} \) and \( z_{t}^{\textcolor{orange}{Attr}} \) as well as between \( z_{t}^{\textcolor{cyan}{ID'}} \) and \( z_{t}^{\textcolor{orange}{Attr'}} \), as illustrated in the rightmost diagram of Figure \ref{fig:principle}).
In the Unet architecture of text-guided diffusion models, conditions are incorporated into the model through cross-attention mechanism based on the following equation:
\begin{align}
\scalebox{0.95}{$ %
\operatorname{Attention}\left(\mathbf{Q}, \mathbf{K}, \mathbf{V}\right) 
= \operatorname{Softmax}\left(\frac{\mathbf{Q}\left(\mathbf{K}\right)^{\top}}{\sqrt{d}}\right) \mathbf{V},
$}
\end{align}
where \( \mathbf{Q} = \varphi(z_{t})W_{Q} \), \( \mathbf{K} = \mathcal{C}W_{K} \), and \( \mathbf{V} = \mathbf{C}W_{V} \) represent the query, key, and value matrices, respectively. \( \varphi(z_{t}) \) indicates the intermediate spatial features of the U-Net. \( W_{Q} \), \( W_{K} \), and \( W_{V} \) are trainable weight matrices, and \( \mathcal{C} = \mathcal{E}_{text}(P) \) represents the text embedding of face description \( P \) through CLIP text encoder.

We insert face image embedding conditions by adding a new cross-attention layer alongside the text cross-attention layer, following the same mechanism as some face-driven image generation works \cite{IP-adapter, InstantID}. The output features of the cross-attention layers \(Z_{\text{out}}\) are computed as follows:
\begin{align}
Z_{\text{out}} = \text{Attention}(\mathbf{Q}, \mathbf{K}, \mathbf{V}) + k \, \text{Attention}(\mathbf{Q}, \mathbf{K'}, \mathbf{V'})
\label{eq:atten}
\end{align}
where \( \mathbf{K'} = \mathcal{C'} W'_{K} \) and \( \mathbf{V'} = \mathcal{C'} W'_{V} \), with \( \mathcal{C'} = \mathcal{F}(\mathcal{E}_{\text{vis}}(I)) \) representing the image embedding of \( I \) through the CLIP vision encoder \(\mathcal{E}_{\text{vis}}(\cdot)\) and projection network \(\mathcal{F}(\cdot) \), \( k \in [0, 1] \) is the scaling factor for controlling the attention intensity of condition \(\mathcal{C'}\), and \( W_{Q}' \), \( W_{K}' \), and \( W_{V}' \) are trainable weight matrices.

We use a pretrained Stable Diffusion model, keeping its parameters fixed while adding LoRA \cite{lora} layers to enable lightweight training. The cross-attention layers, projection model, and LoRA weights are trained based on the following loss function:
\begin{align}
L=\mathbb{E}_{\boldsymbol{z}_0, \boldsymbol{\epsilon} \sim \mathcal{N}(\mathbf{0}, \mathbf{I}), \mathcal{C}, \mathcal{C^{'}} , t}\left\|\boldsymbol{\epsilon}-\boldsymbol{\epsilon}_\theta\left(\boldsymbol{z}_t, \mathcal{C}, \mathcal{C^{'}} , t\right)\right\|^2,
\end{align}

\textbf{Face editing (Inversion).} First, we invert the image to the latent space of the diffusion model. We use the entire face image embedding \( \mathcal{C'} \) and the text description embedding \( \mathcal{C} \) as guiding conditions to perform DDIM inversion, obtaining an initial latent code \( \boldsymbol{z}^{*}_T \). The inversion process is shown below:
\begin{align}
    z_{t+1} = &\ \sqrt{\bar{\alpha}_{t+1}} f_\theta\left(z_t, \mathcal{C}, \mathcal{C^{'}} , t\right) \notag \\
    &+ \sqrt{1 - \bar{\alpha}_{t+1}} \, \epsilon_\theta\left(z_t, \mathcal{C}, \mathcal{C^{'}} , t\right),
\end{align}
where \( f_\theta\left(z_t, \mathcal{C}, t\right) \) represents the model's prediction of \( z_0 \) at each time step, \( \bar{\alpha}_{t} \) is a scaling factor as defined in DDIM \cite{DDIM}.

During inversion process, we set the CFG scale \( \omega = 1 \) to obtain a precise initial latent code that is unaffected by the unconditional component.

\textbf{Face editing (Reverse diffusion process).}

 After inversion, we perform the reverse diffusion process with CFG (\(\omega > 1\)), starting from the initial latent code. In CFG, we use the modified prompt \( \mathcal{C}_n = \mathcal{E}_{\text{text}}(P_n) \) and entire image embedding \(\mathcal{C'}\) as the positive components. The original prompt (i.e., face image text description) and a zero values embedding \(\mathcal{C'}_{\text{zero}}\), matching the shape of \(\mathcal{C'}\), serve as the negative components. The CFG is represented as follows:
\begin{align}
\tilde{\boldsymbol{\epsilon}}_\theta\left(\boldsymbol{z}_t, \mathcal{C}, \mathcal{C}_{n}, \mathcal{C'}, \mathcal{C'}_{\text{zero}}, t\right)
= \boldsymbol{\epsilon}_\theta\left(\boldsymbol{z}_t, \mathcal{C}, \mathcal{C'}_{\text{zero}}, t\right) \notag \\
+ \omega \left( \boldsymbol{\epsilon}_\theta\left(\boldsymbol{z}_t, \mathcal{C}_n, \mathcal{C'}, t\right) 
- \boldsymbol{\epsilon}_\theta\left(\boldsymbol{z}_t, \mathcal{C}, \mathcal{C'}_{\text{zero}}, t\right) \right)
\end{align}

For example, when transforming an image of a man into a woman wearing glasses, the negative prompt during the reverse diffusion process would be ``a man,'' while the positive prompt would be ``a woman wearing glasses.'' Intuitively, this setup enables the model to reduce the influence of the original attribute through CFG, guiding the editing toward the direction of the positive prompt. Additionally, as demonstrated in Negative-Prompt Inversion \cite{negativeprompt}, using the original prompt as the negative prompt in CFG can be regarded as a mathematical approximation of Null-Text Inversion \cite{nulltext}, effectively improving editing structural consistency. 

The scaled face image embedding has only an intensity adjustment by  \( k \), preserving its structural integrity. This condition controls the alignment of latent codes in the reverse diffusion process with those from the inversion, providing a stronger constraint as \( k \to 1 \) in equation \ref{eq:atten}.

\section{Experiment}
\subsection{Experimental setting}
\noindent \textbf{Dataset Construction.} For face attribute-specific training, we created a dataset consisting of 69,900 face image-text description pairs. The original images were sampled from the FFHQ 
\cite{FFHQ} dataset and resized to 512×512 resolution. We use OpenAI's vision API which is based on GPT-4o to generate attribute text description for each face image, capturing details such as expression, body type, hair color, ethnicity, gender, presence of glasses, and facial hair. For specific age information, we incorporated age labels from the FFHQ-Aging dataset \cite{FFHQ-aging} and inserted them into the description text. An example description is: ``A chubby Indian man, aged 20 to 29, with black hair, glasses, and a beard, smiling."

For evaluation, we sampled 100 images from the FFHQ dataset (distinct from those in the training set) and additional 100 real face images randomly selected from the CelebA-HQ \cite{celebA-HQ} dataset.
\begin{table}[t]
\centering
\caption{Face Editing Experiment Tasks}
\label{tab:singandmul}
\resizebox{\linewidth}{!}{%
\begin{tabular}{@{}l|lll@{}}
\toprule
\begin{tabular}[c]{@{}l@{}}Single attribute\\ editing\end{tabular} &
  \multicolumn{3}{l}{\begin{tabular}[c]{@{}l@{}}(1) Expression changes (e.g., smiling, anger, sadness, etc.)\\ (2) Hair color changes (e.g., changing black hair to blonde, pink, or blue)\\ (3) Wearing glasses\\ (4) Age changes (from young to old or vice versa)\\ (5) Gender changes (from male to female or vice versa)\\ (6) Becoming chubby\end{tabular}} \\ \midrule
\begin{tabular}[c]{@{}l@{}}Multi-attribute \\ editing\end{tabular} &
  \multicolumn{3}{l}{\begin{tabular}[c]{@{}l@{}}(1) Age change + wearing glasses\\ (2) Hair color change + gender change\\ (3) Becoming chubby + changing eyes color\\ (4) Changing ethnicity\end{tabular}} \\ \bottomrule
\end{tabular}%
}
\end{table}

\noindent \textbf{Task.} We conduct reconstruction experiments, followed by face editing experiments. For face editing experiments, we select six single-attribute editing tasks and four multi-attribute editing tasks, as detailed in Table \ref{tab:singandmul}.

\noindent \textbf{Baseline Methods.} We compare our method with several state-of-the-art face editing methods, with a primary focus on diffusion model-based methods. For GAN-based methods, we adopt StyleClip \cite{StyleClip}, which is driven by text prompts and supports multi-attribute editing. Among diffusion model-based methods, we utilize Diffusion Autoencoder \cite{DiffusionAutoencoders} and Null-text Inversion \cite{nulltext}, fine-tuned on the FFHQ dataset, as well as the recently proposed Collaborative Diffusion \cite{collaborative}, which is specifically designed for face generation and editing tasks.

\noindent \textbf{Metrics.} We use mean squared error (MSE), structural similarity (SSIM), and peak signal-to-noise ratio (PSNR) to evaluate the quality of the reconstructed images. For editing tasks, we employ Structure Dist \cite{StructureDist} to evaluate structural consistency, where lower values indicate that the edited image is more similar to the original. Additionally, we use ID similarity (ID) to evaluate the ID consistency, which is calculated as the cosine similarity between the feature vectors of the original image and the edited image, extracted by a pre-trained face recognition model. Furthermore, we trained a face attribute recognition model and calculate its accuracy (Acc) to quantitatively evaluate whether the specified editing target is achieved. Finally, we utilize the no-reference image quality assessment metric BIRQUSE \cite{BIRSQUE} to evaluate the quality of the edited images.

\subsection{Results}
\begin{figure}[t]
  \centering
  \includegraphics[width=\linewidth]{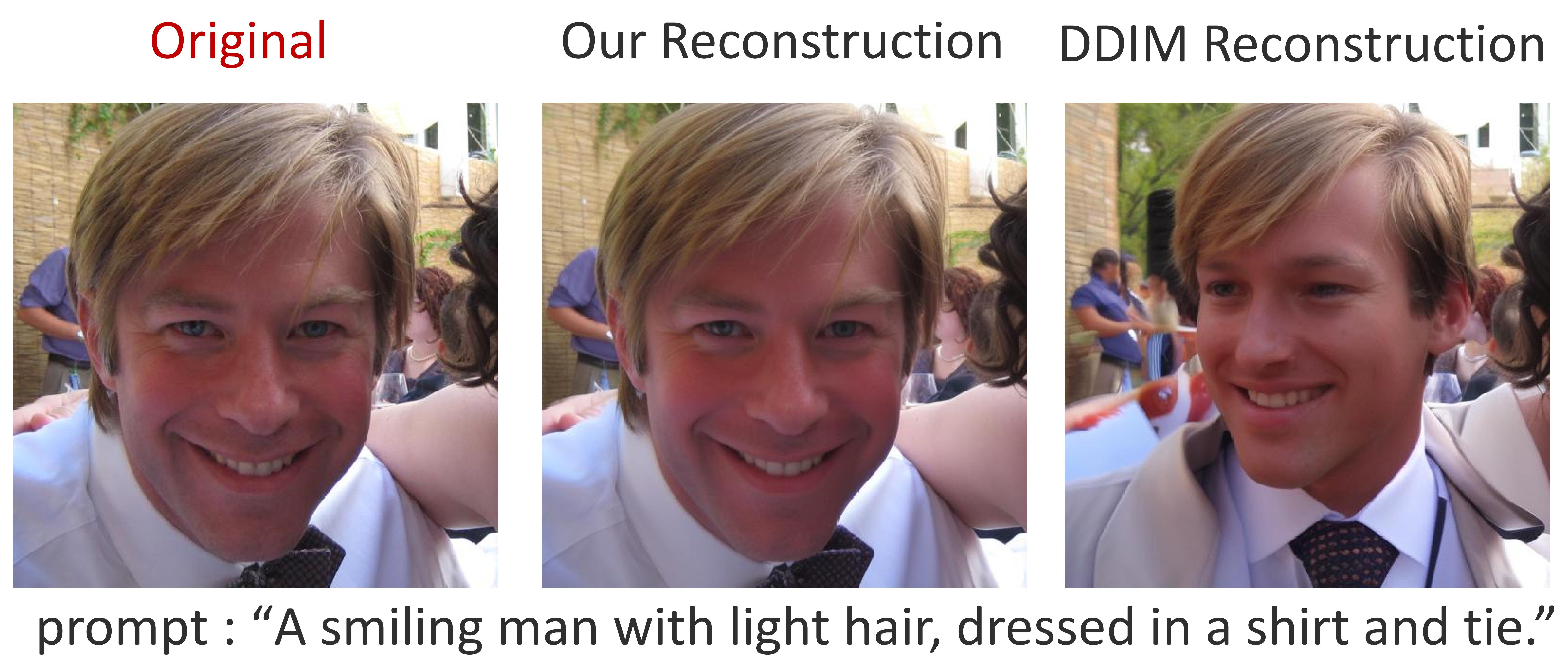}
  \caption{Comparison of reconstruction results between our method and text-guided DDIM inversion.}
  \label{fig:recon}
\end{figure}

\begin{table}[t]
\centering
\caption{A Quantitative Comparison Between Our Method and Text-Guided Inversion for Reconstruction.}
\label{tab:recon}
\resizebox{0.8\linewidth}{!}{%
\begin{tabular}{@{}l|c|c|c@{}}
\toprule
Methods    & MSE $\downarrow$            & SSIM $\uparrow$          & PSNR  $\uparrow$                               \\ \midrule
Text-guided DDIM & 70.312          & 0.587          & \multicolumn{1}{c}{27.032}          \\ \midrule
Ours reconstruction & \textbf{22.051} & \textbf{0.878} & \multicolumn{1}{c}{\textbf{34.064}} \\ \bottomrule
\end{tabular}%
}
\end{table}
Due to page limitations, we present only a subset of the editing result comparisons in the main text; for the complete set, please refer to the Appendix.

Figure \ref{fig:recon} compares the reconstruction results of our method with text-guided DDIM inversion. As shown, even with detailed and comprehensive text descriptions, text-guided DDIM inversion fails to precisely reconstruct the original images and introduces noticeable artifacts in some cases. In contrast, our method achieves highly detailed and stable reconstructions, capturing fine features such as hair and beards. It is important to note that minor detail loss is an inherent limitation of using Stable Diffusion, as the image encoder tends to slightly smooth the original input. However, this loss remains negligible and does not noticeably impact the visual quality of the reconstructions. Table \ref{tab:recon} quantitatively presents the reconstruction results, demonstrating that our method significantly outperforms the text-guided DDIM inversion.
\begin{figure*}[t]
  \centering
  \includegraphics[width=\textwidth]{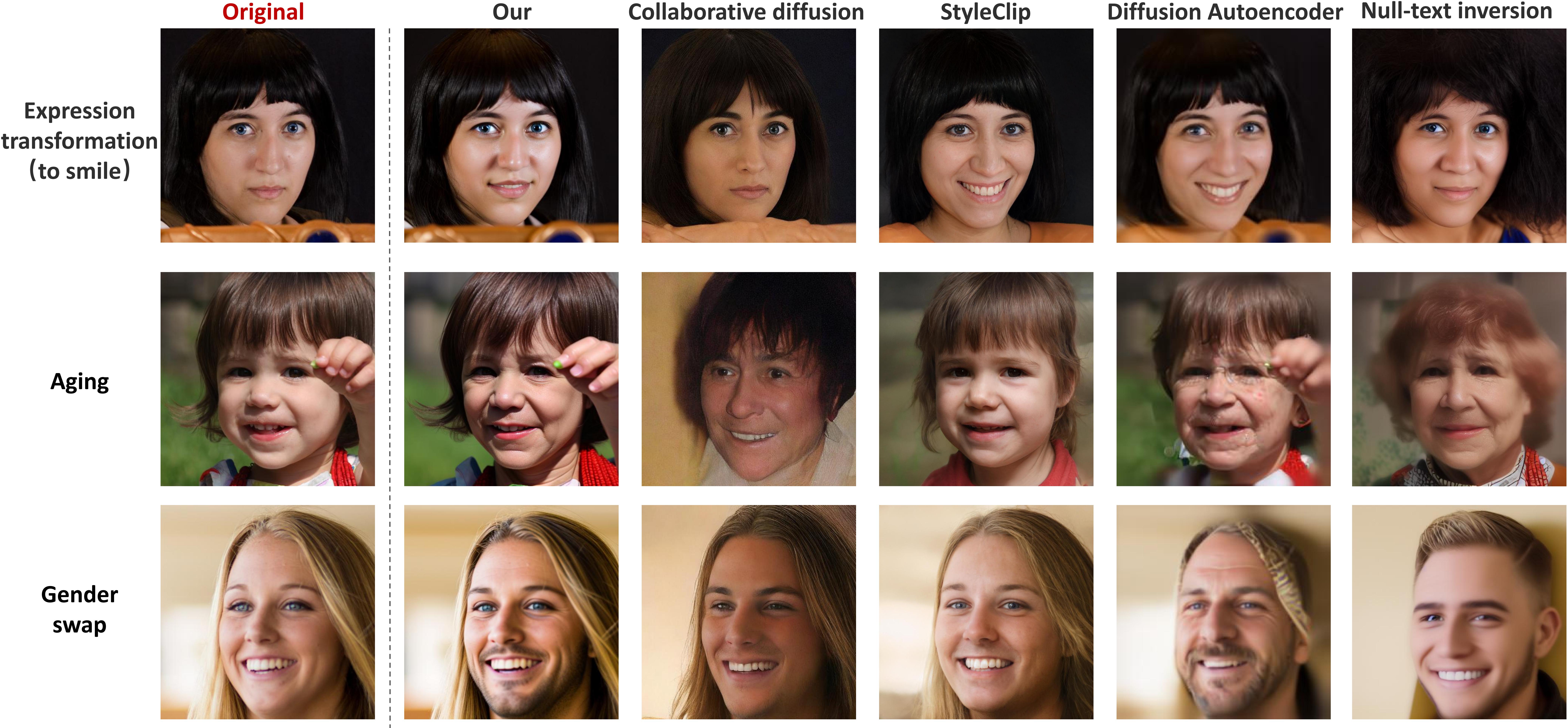}
  \caption{Comparison of different methods on single-attribute editing tasks. Each row corresponds to a different attribute editing task. It can be seen that our method outperforms existing approaches in terms of editing accuracy, as well as ID and structural consistency. (Zoom in to see details.)}
  \label{fig:comparison}
\end{figure*}
\begin{table}[t]
\centering
\caption{A quantitative comparison of our method with SOTA face editing methods on single-attribute editing tasks.}
\label{tab:single}
\resizebox{\linewidth}{!}{%
\begin{tabular}{@{}l|c|c|c|c@{}}
\toprule
Methods   & \begin{tabular}[c]{@{}c@{}}Struct Dist $\downarrow$ \end{tabular} & ID  $\uparrow$           & Acc $\uparrow$             & BIRSQUE $\downarrow$       \\ \midrule
StyleClip \cite{StyleClip} & 0.042                                                     & 0.804          & 80.42\%          & 35.28          \\
DiffAE \cite{DiffusionAutoencoders}    & 0.047                                                     & 0.851         & 82.23\%         & 40.45          \\
Collab \cite{collaborative}    & 0.060                                                     & 0.301          & 29.15\%          & 48.31          \\
Null-text \cite{nulltext} & 0.058                                                     & 0.562          & 73.21\%          & 56.26          \\ \midrule
Ours      & \textbf{0.025}                                            & \textbf{0.884} & \textbf{84.62} & \textbf{27.63} \\ \bottomrule
\end{tabular}%
}
\end{table}

\vspace{0.5cm}
\noindent \textbf{Single-attribute editing results.} As shown in Figure \ref{fig:comparison}, our method accurately edits target features while maintaining both ID and structural consistency. It also preserves non-facial details, such as hands in aging tasks and hair in gender transformation tasks. Collaborative Diffusion achieves semantically valid edits in specific tasks but fails to maintain ID and structural consistency, as it relies on semantic masks and requires extensive fine-tuning per image. StyleCLIP performs target edits but significantly alters the ID, structure, and background. Diffusion Autoencoder preserves ID and structure to some extent but often produces blurred backgrounds and artifacts. Null-text inversion fails to maintain ID or structural consistency, highlighting the limitations of general image editing methods in face editing tasks.

Table \ref{tab:single} quantitatively compares the methods, showing that our approach outperforms others in ID consistency, structural consistency, editing accuracy, and image quality.

\begin{figure}[t]
  \centering
  \includegraphics[width=\linewidth]{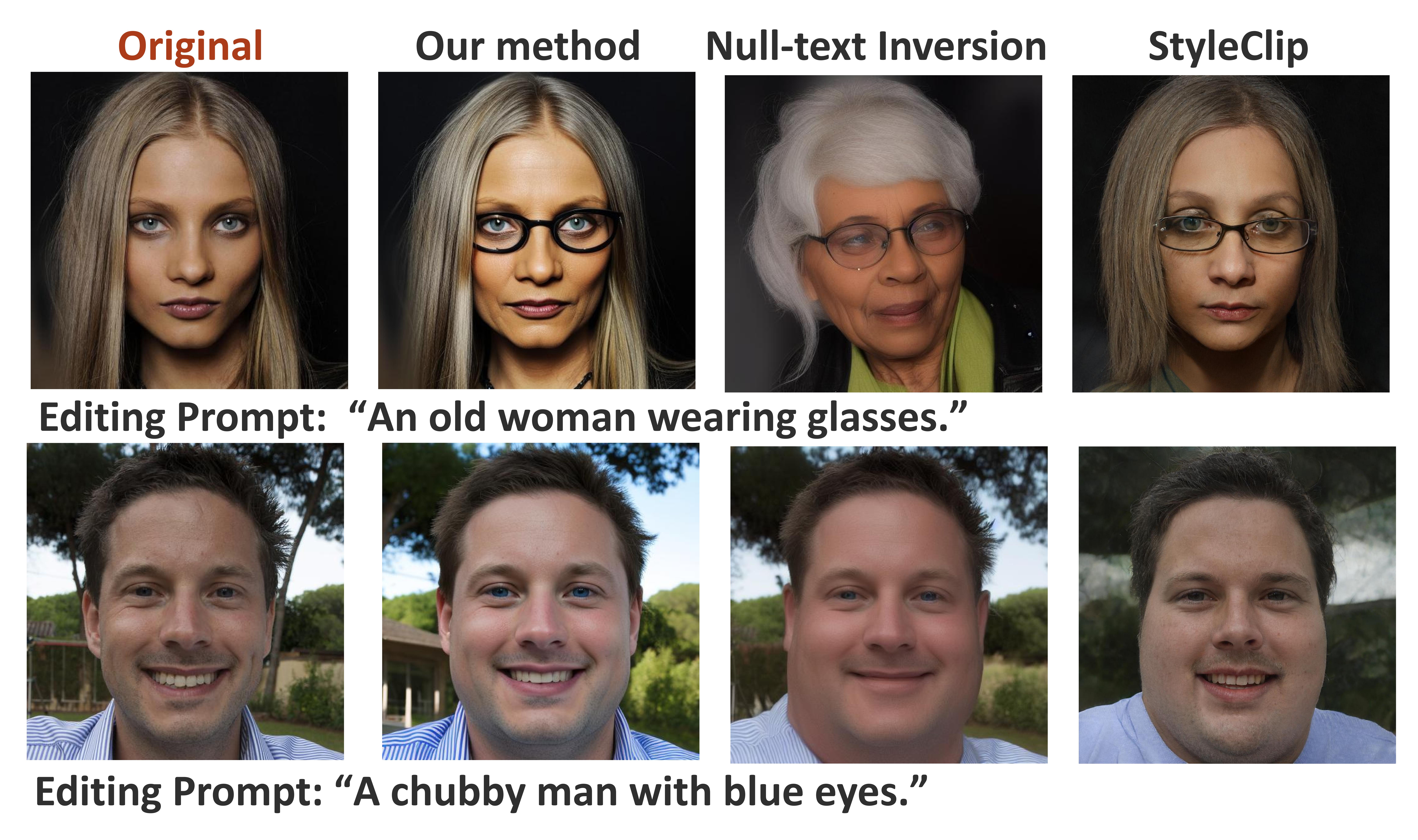}
  \caption{Comparison of different methods on multi-attribute editing tasks. It can be seen that our method still achieves high-quality editing results in multi-attribute editing tasks, maintaining both ID and structural consistency.}
  \label{fig:multi}
\end{figure}

\begin{table}[t]
\centering
\caption{A quantitative comparison of our method with SOTA face editing methods on multi-attribute editing tasks.}
\label{tab:multi}
\resizebox{\linewidth}{!}{%
\begin{tabular}{@{}l|c|c|c|c@{}}
\toprule
Methods   & \begin{tabular}[c]{@{}c@{}}Struct Dist $\downarrow$ \end{tabular} & ID $\uparrow$           & Acc $\uparrow$             & BIRSQUE $\downarrow$       \\ \midrule
StyleClip \cite{StyleClip} & 0.060                                                     & 0.62          & 60.37\%          & 38.45          \\
Null-text \cite{nulltext} & 0.053                                                     & 0.48          & 73.25\%          & 53.75          \\ \midrule
Ours      & \textbf{0.035}                                            & \textbf{0.79} & \textbf{75.32\%} & \textbf{28.31} \\ \bottomrule
\end{tabular}%
}
\end{table}

\noindent \textbf{Multi-attribute editing results.} Multi-attribute editing requires prompts specifying multiple attribute transformations simultaneously. Diffusion AutoEncoder only supports single-attribute editing, while Collaborative Diffusion requires semantic masks, which are difficult to provide accurately before editing. Therefore, for fair comparisons, we only compare our method with StyleCLIP and Null-text Inversion, as both methods enable editing using complex prompts as input. As shown in Figure \ref{fig:multi}, our method still achieves high-quality editing results in multi-attribute editing tasks, maintaining both ID and structural consistency. In contrast, StyleClip performs worse than in single-attribute editing, indicating its difficulty in effectively disentangling and manipulating multiple target attributes. Null-text Inversion struggles to maintain detail consistency, resulting in overall lower quality in the edited images. Table \ref{tab:multi} presents the quantitative evaluation results on multi-attribute editing tasks, demonstrating that our method still outperforms other methods.

\section{Conclusion}
In this paper, we propose a zero-shot face editing method based on ID-Attribute Decoupled Inversion, which supports a wide range of complex face editing tasks while maintaining ID and structural consistency. The core idea of our method is to decouple ID features and attribute features through conditional inputs, enabling independent control over both. Our study provides an important insight that high-dimensional structured face image embeddings can serve as precise and robust conditions to constrain the diffusion trajectory, thereby ensuring structural consistency in edited images. This approach is simple yet effective, requiring no complex structural design, and successfully overcomes the bottleneck of current inversion-based image editing methods, which struggle to handle face editing tasks. We conduct extensive comparative experiments, providing both quantitative and qualitative analyses. Experimental results demonstrate that our method outperforms existing approaches in terms of editing accuracy as well as the preservation of ID and structural consistency. 

\bibliographystyle{IEEEbib}
\bibliography{reference}

\begin{thebibliography}{10}

\bibitem{shen2020interpreting}
Yujun Shen, Jinjin Gu, Xiaoou Tang, and Bolei Zhou,
\newblock ``Interpreting the latent space of gans for semantic face editing,''
\newblock in {\em Proceedings of the IEEE/CVF conference on computer vision and pattern recognition}, 2020, pp. 9243--9252.

\bibitem{stablediffusion}
Robin Rombach, Andreas Blattmann, Dominik Lorenz, Patrick Esser, and Bj{\"o}rn Ommer,
\newblock ``High-resolution image synthesis with latent diffusion models,''
\newblock in {\em Proceedings of the IEEE/CVF conference on computer vision and pattern recognition}, 2022, pp. 10684--10695.

\bibitem{Glide}
Alex Nichol, Prafulla Dhariwal, Aditya Ramesh, Pranav Shyam, Pamela Mishkin, Bob McGrew, Ilya Sutskever, and Mark Chen,
\newblock ``Glide: Towards photorealistic image generation and editing with text-guided diffusion models,''
\newblock {\em arXiv preprint arXiv:2112.10741}, 2021.

\bibitem{ramesh2022hierarchical}
Aditya Ramesh, Prafulla Dhariwal, Alex Nichol, Casey Chu, and Mark Chen,
\newblock ``Hierarchical text-conditional image generation with clip latents,''
\newblock {\em arXiv preprint arXiv:2204.06125}, vol. 1, no. 2, pp. 3, 2022.

\bibitem{ho2022classifier}
Jonathan Ho and Tim Salimans,
\newblock ``Classifier-free diffusion guidance,''
\newblock {\em arXiv preprint arXiv:2207.12598}, 2022.

\bibitem{plug-and-play}
Narek Tumanyan, Michal Geyer, Shai Bagon, and Tali Dekel,
\newblock ``Plug-and-play diffusion features for text-driven image-to-image translation,''
\newblock in {\em Proceedings of the IEEE/CVF Conference on Computer Vision and Pattern Recognition}, 2023, pp. 1921--1930.

\bibitem{pix2pixzero}
Gaurav Parmar, Krishna Kumar~Singh, Richard Zhang, Yijun Li, Jingwan Lu, and Jun-Yan Zhu,
\newblock ``Zero-shot image-to-image translation,''
\newblock in {\em ACM SIGGRAPH 2023 Conference Proceedings}, 2023, pp. 1--11.

\bibitem{DDIM}
Jiaming Song, Chenlin Meng, and Stefano Ermon,
\newblock ``Denoising diffusion implicit models,''
\newblock {\em arXiv preprint arXiv:2010.02502}, 2020.

\bibitem{sdedit}
Chenlin Meng, Yutong He, Yang Song, Jiaming Song, Jiajun Wu, Jun-Yan Zhu, and Stefano Ermon,
\newblock ``Sdedit: Guided image synthesis and editing with stochastic differential equations,''
\newblock {\em arXiv preprint arXiv:2108.01073}, 2021.

\bibitem{DiffusionAutoencoders}
Konpat Preechakul, Nattanat Chatthee, Suttisak Wizadwongsa, and Supasorn Suwajanakorn,
\newblock ``Diffusion autoencoders: Toward a meaningful and decodable representation,''
\newblock in {\em Proceedings of the IEEE/CVF conference on computer vision and pattern recognition}, 2022, pp. 10619--10629.

\bibitem{collaborative}
Ziqi Huang, Kelvin~CK Chan, Yuming Jiang, and Ziwei Liu,
\newblock ``Collaborative diffusion for multi-modal face generation and editing,''
\newblock in {\em Proceedings of the IEEE/CVF Conference on Computer Vision and Pattern Recognition}, 2023, pp. 6080--6090.

\bibitem{IP-adapter}
Hu~Ye, Jun Zhang, Sibo Liu, Xiao Han, and Wei Yang,
\newblock ``Ip-adapter: Text compatible image prompt adapter for text-to-image diffusion models,''
\newblock {\em arXiv preprint arXiv:2308.06721}, 2023.

\bibitem{InstantID}
Qixun Wang, Xu~Bai, Haofan Wang, Zekui Qin, Anthony Chen, Huaxia Li, Xu~Tang, and Yao Hu,
\newblock ``Instantid: Zero-shot identity-preserving generation in seconds,''
\newblock {\em arXiv preprint arXiv:2401.07519}, 2024.

\bibitem{photomaker}
Zhen Li, Mingdeng Cao, Xintao Wang, Zhongang Qi, Ming-Ming Cheng, and Ying Shan,
\newblock ``Photomaker: Customizing realistic human photos via stacked id embedding,''
\newblock in {\em Proceedings of the IEEE/CVF Conference on Computer Vision and Pattern Recognition}, 2024, pp. 8640--8650.

\bibitem{clip}
Alec Radford, Jong~Wook Kim, Chris Hallacy, Aditya Ramesh, Gabriel Goh, Sandhini Agarwal, Girish Sastry, Amanda Askell, Pamela Mishkin, Jack Clark, et~al.,
\newblock ``Learning transferable visual models from natural language supervision,''
\newblock in {\em International conference on machine learning}. PMLR, 2021, pp. 8748--8763.

\bibitem{lora}
Edward~J Hu, Yelong Shen, Phillip Wallis, Zeyuan Allen-Zhu, Yuanzhi Li, Shean Wang, Lu~Wang, and Weizhu Chen,
\newblock ``Lora: Low-rank adaptation of large language models,''
\newblock {\em arXiv preprint arXiv:2106.09685}, 2021.

\bibitem{negativeprompt}
Daiki Miyake, Akihiro Iohara, Yu~Saito, and Toshiyuki Tanaka,
\newblock ``Negative-prompt inversion: Fast image inversion for editing with text-guided diffusion models,''
\newblock {\em arXiv preprint arXiv:2305.16807}, 2023.

\bibitem{nulltext}
Ron Mokady, Amir Hertz, Kfir Aberman, Yael Pritch, and Daniel Cohen-Or,
\newblock ``Null-text inversion for editing real images using guided diffusion models,''
\newblock in {\em Proceedings of the IEEE/CVF Conference on Computer Vision and Pattern Recognition}, 2023, pp. 6038--6047.

\bibitem{FFHQ}
Tero Karras, Samuli Laine, and Timo Aila,
\newblock ``A style-based generator architecture for generative adversarial networks,''
\newblock in {\em Proceedings of the IEEE/CVF conference on computer vision and pattern recognition}, 2019, pp. 4401--4410.

\bibitem{FFHQ-aging}
Roy Or-El, Soumyadip Sengupta, Ohad Fried, Eli Shechtman, and Ira Kemelmacher-Shlizerman,
\newblock ``Lifespan age transformation synthesis,''
\newblock in {\em Computer Vision--ECCV 2020: 16th European Conference, Glasgow, UK, August 23--28, 2020, Proceedings, Part VI 16}. Springer, 2020, pp. 739--755.

\bibitem{celebA-HQ}
Tero Karras,
\newblock ``Progressive growing of gans for improved quality, stability, and variation,''
\newblock {\em arXiv preprint arXiv:1710.10196}, 2017.

\bibitem{StyleClip}
Or~Patashnik, Zongze Wu, Eli Shechtman, Daniel Cohen-Or, and Dani Lischinski,
\newblock ``Styleclip: Text-driven manipulation of stylegan imagery,''
\newblock in {\em Proceedings of the IEEE/CVF international conference on computer vision}, 2021, pp. 2085--2094.

\bibitem{StructureDist}
Narek Tumanyan, Omer Bar-Tal, Shai Bagon, and Tali Dekel,
\newblock ``Splicing vit features for semantic appearance transfer,''
\newblock in {\em Proceedings of the IEEE/CVF Conference on Computer Vision and Pattern Recognition}, 2022, pp. 10748--10757.

\bibitem{BIRSQUE}
Anish Mittal, Anush~Krishna Moorthy, and Alan~Conrad Bovik,
\newblock ``No-reference image quality assessment in the spatial domain,''
\newblock {\em IEEE Transactions on image processing}, vol. 21, no. 12, pp. 4695--4708, 2012.

\end{thebibliography}

\end{document}